\documentclass{article}

\usepackage[preprint]{neurips_2025}
\usepackage{fancyhdr} 
\usepackage{blindtext} 
\usepackage{makecell}
\usepackage{natbib}

\usepackage{fontawesome5} 

\usepackage[utf8]{inputenc} 
\usepackage[T1]{fontenc}    
\usepackage{hyperref}       
\usepackage{url}            
\usepackage{booktabs}       
\usepackage{amsfonts}       
\usepackage{nicefrac}       

\usepackage{xcolor}         
\usepackage{amsmath}
\usepackage{float}
\usepackage{tabularray}
\usepackage{multirow}
\usepackage{booktabs, caption, array}
\usepackage{amssymb}
\usepackage{lipsum}         
\usepackage{CJKutf8}        
\usepackage{nomencl}
\usepackage{glossaries}
\usepackage{xspace}
\usepackage{multirow}
\usepackage{graphicx}
\usepackage{bbding}
\usepackage{pifont}
\usepackage{algorithm}
\usepackage{enumitem}
\usepackage{multicol}
\usepackage{caption}
\usepackage{array}
\usepackage{fp}
\usepackage{multirow}   
\usepackage{makecell}
\usepackage{flushend}
\usepackage{cases}
\usepackage{color}
\usepackage{algpseudocode}
\usepackage{listings}
\usepackage{mathrsfs}
\usepackage{longtable}
\usepackage{colortbl}
\usepackage{bm}
\usepackage{caption}
\usepackage{natbib}
\usepackage{tabularx}
\usepackage{xcolor}         
\usepackage{booktabs}    
\usepackage{multirow}   
\usepackage{graphicx}    
\usepackage[table]{xcolor} 
\usepackage{enumitem}
\usepackage{amsmath}
\usepackage{svg}
\usepackage{xurl}   
\usepackage{hyperref}
\usepackage{CJKutf8}
\definecolor{softblue}{rgb}{0.8, 0.9, 1.0}
\definecolor{softgreen}{rgb}{0.0, 0.6, 0.2}

\usepackage{subcaption} 

\setlist[itemize]{noitemsep, topsep=0pt}
\usepackage{arydshln}
\usepackage{lipsum}
\definecolor{codegreen}{rgb}{0,0.3,0.6}
\definecolor{codegray}{rgb}{0.5,0.5,0.5}

\newcommand{\ie}{\emph{i.e.}\xspace}
\newcommand{\eg}{\emph{e.g.}\xspace}

\newcommand{\ignore}[1]{}

\usepackage[most]{tcolorbox} 
\usepackage{microtype}      
\definecolor{darkorange}{RGB}{255, 140, 0}
\definecolor{lightgreen}{RGB}{145, 204, 117}
\definecolor{lightyellow}{RGB}{250, 200, 88}
\definecolor{lightred}{RGB}{238, 102, 102}
\definecolor{lightblue}{RGB}{115, 192, 222}
\newtcolorbox{promptbox}[3][Judge Prompt]{
colback=black!5!white,
arc=5pt, 
boxrule=0.5pt,
fonttitle=\bfseries,
title=#1, 
before upper={\small}, fontupper=\fontfamily{ptm}\selectfont,
colframe=#2,
label=#3,
}
\definecolor{gray_1}{HTML}{B7B7B7}
\definecolor{gray_2}{HTML}{F0F0F0} 
\definecolor{frame_blue}{HTML}{A9D18E}

\newtcolorbox[auto counter, number within=section]{PromptBoxNew}[2][]{
    enhanced,
    breakable,
    colback=gray_2, 
    colframe=gray_1,
    coltitle=white,
    fontupper=\small,
    fonttitle=\bfseries,
    title={#2}, 
    label={#1},
    arc=2pt,
    boxrule=1pt,
    left=2mm, right=2mm, top=2mm, bottom=2mm,
}

\usepackage{etoolbox}
\makeatletter
\patchcmd{\@maketitle}
  {\@toptitlebar}
  {%
    \hbox to \textwidth{%
    
    \includegraphics[height=0.76cm]{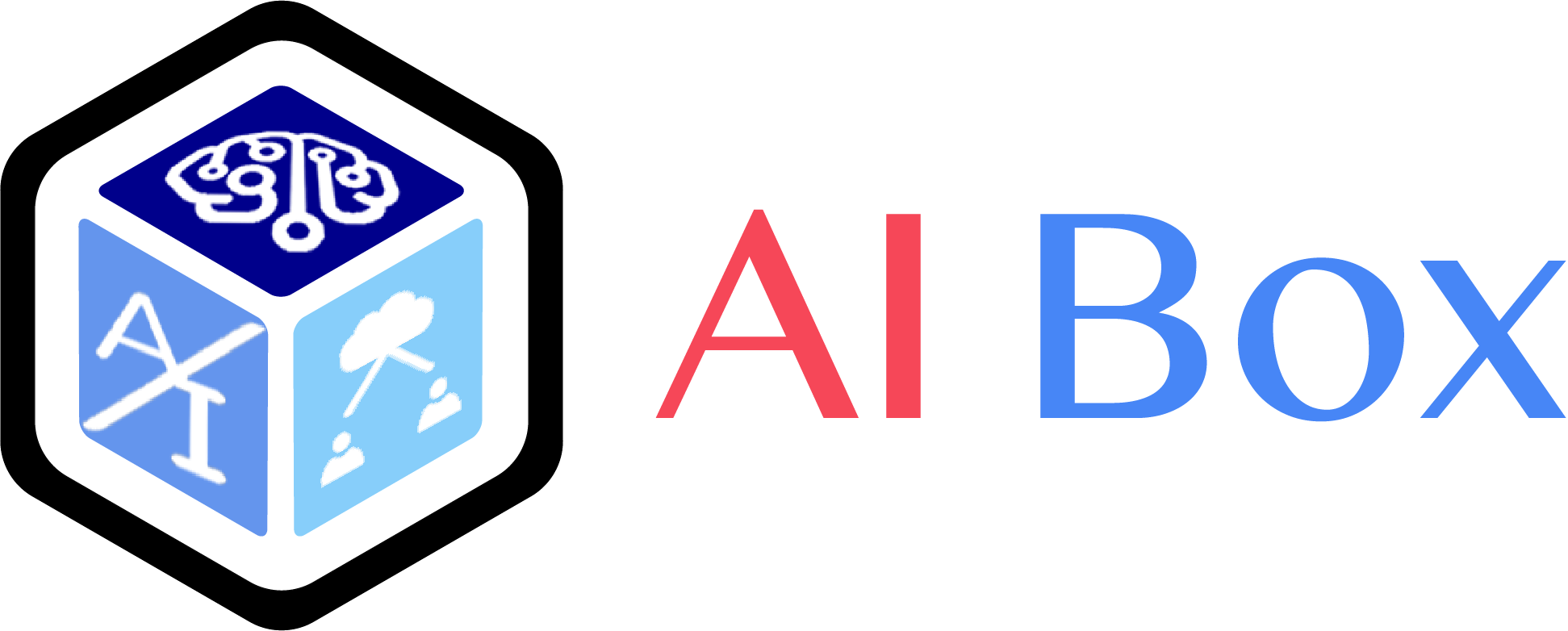}%
      \hfill
    }%
    \vspace{0.3em}
    \@toptitlebar
  }
  {}{}
\makeatother

\title{Towards Long-horizon Agentic Multimodal Search}

\author{%
    Yifan Du$^{1}$\thanks{Equal contribution.}~, 
  Zikang Liu$^{1*}$, 
  Jinbiao Peng$^{1}$, 
  Jie Wu$^{1}$,
  Junyi Li$^2$,
  \textbf{Jinyang Li$^{1}$},\\
  \textbf{Wayne Xin Zhao$^{1}$\thanks{Corresponding author}~,}  \textbf{Ji-Rong Wen}$^{1}$
  \\
  $^1$Gaoling School of Artificial Intelligence, Renmin University of China.\\
  $^2$City University of Hong Kong.\\
  \texttt{yifandu1999@gmail.com},
  \texttt{batmanfly@gmail.com}
}

\begin{document}
\maketitle
\setcounter{footnote}{0}



\definecolor{babyblue}{HTML}{F8F9FE}
\newtcolorbox{bluebox}{
  colback=babyblue,    
  colframe=babyblue,  
  width=1.0\textwidth,  
  center,               
  arc=8pt,                 
  boxrule=0pt,           
  boxsep=0pt,       
  left=2pt,               
  right=2pt,              
  top=10pt,                
  bottom=10pt              
}


\begin{bluebox}
\begin{abstract}

Multimodal deep search agents have shown great potential in solving complex tasks by iteratively collecting textual and visual evidence. However, managing the heterogeneous information and high token costs associated with multimodal inputs over long horizons remains a critical challenge, as existing methods often suffer from context explosion or the loss of crucial visual signals. To address this, we propose a novel \textbf{L}ong-horizon \textbf{M}ulti\textbf{M}odal deep search framework, named \textbf{LMM-Searcher}, centered on a file-based visual representation mechanism. By offloading visual assets to an external file system and mapping them to lightweight textual identifiers (UIDs), our approach mitigates context overhead while preserving multimodal information for future access. We equip the agent with a tailored \emph{fetch-image} tool, enabling a progressive, on-demand visual loading strategy for active perception. Furthermore, we introduce a data synthesis pipeline designed to generate queries requiring complex cross-modal multi-hop reasoning. Using this pipeline, we distill 12K high-quality trajectories to fine-tune Qwen3-VL-Thinking-30A3B into a specialized multimodal deep search agent. Extensive experiments across four benchmarks demonstrate that our method successfully scales to 100-turn search horizons, achieving state-of-the-art performance among open-source models on challenging long-horizon benchmarks like MM-BrowseComp and MMSearch-Plus, while also exhibiting strong generalizability across different base models. Our code will be released in \url{https://github.com/RUCAIBox/LMM-Searcher}.
\end{abstract}
\end{bluebox}

\section{Introduction}
Deep search agent systems~\cite{jin2025search,song2025r1,li2025search} have achieved significant success in tackling challenging real-world information-seeking problems~\cite{wei2025browsecomp,mialon2023gaia}. Building on the deep search framework, these systems can query search engines and browse web pages to iteratively gather factual evidence, thereby solving complex tasks. A key distinction from traditional search systems is that deep search systems often engage in a long-horizon process of iterative reasoning and evidence accumulation, progressively working toward the final solution to a given problem. Recent work has extended this paradigm to multimodal search agents~\cite{geng2025webwatcher,huang2026vision,chu2026redsearcher} by incorporating specialized visual tools such as image search.

However, the multimodal search process~\cite{mei2025survey} differs significantly from purely language-based search. The information gathered during searching and browsing is heterogeneous~\cite{du2022survey,yin2024survey} and suffers from context explosion due to the high token cost of multimodal inputs (\eg, images or videos)~\cite{yao2026towards,wen2025token}. This issue becomes more severe for long-horizon tasks with numerous interactions. 
Prior context management methods that focus on condensing and summarizing textual context histories do not transfer well to deep multimodal search~\cite{mei2025surveycontext,wu2025resum,chen2025iterresearch}. Unlike text, multimodal inputs have fundamentally different data formats and representations~\cite{liang2022mind,li2024multimodal,du2024exploring,shu2025large} and thus cannot be simply treated as a text compression problem. 
In practice, heuristic approaches~\cite{hong2025deepeyesv2} are often adopted to process search results by discarding intermediate image data. Nevertheless, such strategies may cause the loss of important signals, compromising information completeness and making it difficult to scale to long-horizon deep search scenarios.  This raises a central question: \emph{How can we effectively process and manage the accumulated multimodal contexts in the deep search process?}

Inspired by recent progress in the planning-with-files      paradigm~\cite{othmanadi2024planningwithfiles,merrill2026terminal}, we propose to offload multimodal information from the context and store it as external files. In this way, these files can be adaptively loaded, analyzed, or further manipulated progressively during search and reasoning. Such an approach preserves complete multimodal information for future access while reducing context overhead through on-demand loading.

To implement this idea, we propose a long-horizon multimodal context management method centered on a file-based context representation mechanism, named \textbf{LMM-Searcher}. Specifically, all visual assets—whether retrieved from web documents or generated by the environment—are stored in an external file system and mapped to unique textual identifiers (UIDs), which can be further complemented with summary semantics from compact thumbnails. Through these textual proxies, the agent can track multimodal information over long horizons with minimal context cost. To fully leverage this representation, we redesign conventional multimodal search tools and equip the agent with a new tailored tool, \emph{fetch-image}, for active perception. Based on these designs, we develop a progressive multimodal search workflow that allows the agent to retrieve and load specific visual content only when fine-grained understanding is required.

Furthermore, to enhance the agent’s ability in long-horizon multimodal search, we develop a data synthesis pipeline that constructs queries requiring complex cross-modal multi-hop reasoning. Combined with open-source deep search queries, these synthesized tasks are used to collect high-quality trajectories from a strong teacher model for agentic training. Based on this training data, we fine-tune Qwen3-VL-Thinking-30A3B into a specialized multimodal deep search agent.

To validate our approach, we conduct extensive experiments on four multimodal search benchmarks. On challenging long-horizon benchmarks, MM-BrowseComp (MMBC)~\cite{li2025mm} and MMSearch-Plus~\cite{tao2025mmsearch}, our method achieves success rates of 22.3 and 32.9, respectively. Equipped with the context management strategy, our model can scale to 100 turns and achieve performances of 30.1 and 34.8, establishing state-of-the-art results among open-source models. Besides the trained models, our approach also demonstrates superiority in enhancing the model's agentic search capabilities. Specifically, when applied to the same base models, our framework significantly outperforms the prior framework~\cite{chu2026redsearcher}. Based on Seed-1.8, we achieve 46.7 on MMSearch-Plus, demonstrating the strong generalizability of our approach.

Our contributions are summarized below:

\begin{itemize}[leftmargin=1em]
    \item \textbf{Long-horizon multimodal deep search framework:} We propose a novel framework based on file-based visual representation and a specialized agentic tool interface. By offloading visual assets to an external file system and fetching them on demand, our method mitigates the problem of context explosion and scales effectively.

    \item \textbf{Data synthesis pipeline for long-horizon search:} We design a data synthesis pipeline for complex cross-modal multi-hop reasoning. Statistical analysis shows that our synthesized queries require more tool-use turns and involve a higher proportion of vision-related tools than existing datasets.
    
    \item \textbf{A long-horizon multimodal deep search agent:} Based on the framework and the synthesized data, we distill 12K high-quality agent trajectories and fine-tune Qwen3-VL-30A3B-Thinking into a multimodal deep search agent. Extensive experiments across four benchmarks show that our method scales to 100 turns and achieves state-of-the-art performance among open-source models.

\end{itemize}
\section{Related Work}


\paragraph{Language-based Deep Search Agent.}

Language-based deep search agents aim to tackle the inherent limitations of knowledge boundaries for large language models~(LLMs)~\cite{brown2020language,zhao2023survey,liu2024deepseek} by introducing external search and retrieval mechanisms~\cite{lewis2020retrieval,gao2023retrieval,xi2025survey}. Early research typically adopts the retrieval-augmented generation~(RAG) paradigm~\cite{guu2020retrieval,asai2023self,jeong2024adaptive,fan2024survey} to achieve precise knowledge enhancement by retrieving relevant document snippets from static databases via embedding-based methods~\cite{karpukhin2020dense,reimers2019sentence}. Subsequent studies overcome the constraints of pre-built knowledge bases by equipping models with search tools~\cite{li2025search,sun2025simpledeepsearcher,song2025r1}. This integration directly grants models internet search capabilities and further improves their performance in open-domain question answering~\cite{wei2025browsecomp,chen2025browsecomp}. 
However, these language-based agents only support textual search inputs and feedback, limiting their capacity to resolve multimodal queries in real-world applications

\paragraph{Multimodal Deep Search Agent.}
Similar to LLMs, MLLMs~\cite{liu2023visual,yin2024survey,bai2025qwen3} also require external tools to handle complex real-world tasks. Early research~\cite{wu2023visual,liu2024llava} equips models with extensive visual and linguistic plugins, including plugins for object detection~\cite{zou2023object}, image segmentation~\cite{long2015fully}, and OCR~\cite{long2021scene}. This setup enables MLLMs to autonomously invoke appropriate tools based on complex user instructions. Beyond this basic paradigm, recent studies internalize such interactive capabilities into the model's reasoning process, leading to the thinking-with-image paradigm~\cite{du2025virgo,openai2025thinking,su2025thinking,zheng2025deepeyes}. Such frameworks treat visual operations as explicit reasoning steps, facilitating significant gains in spatial reasoning and fine-grained VQA. Building upon these advancements, recent work~\cite{geng2025webwatcher,hong2025deepeyesv2,huang2026vision,chu2026redsearcher} deeply integrates search engines as core tools into the reasoning chain of MLLMs. By combining robust internal visual reasoning with dynamic external search tools, models are empowered to perform complex fact-checking and open-domain multimodal exploration.
\section{Long-horizon Multimodal Context Management}

\begin{figure}[t]
    \centering
    \includegraphics[width=0.9\textwidth]{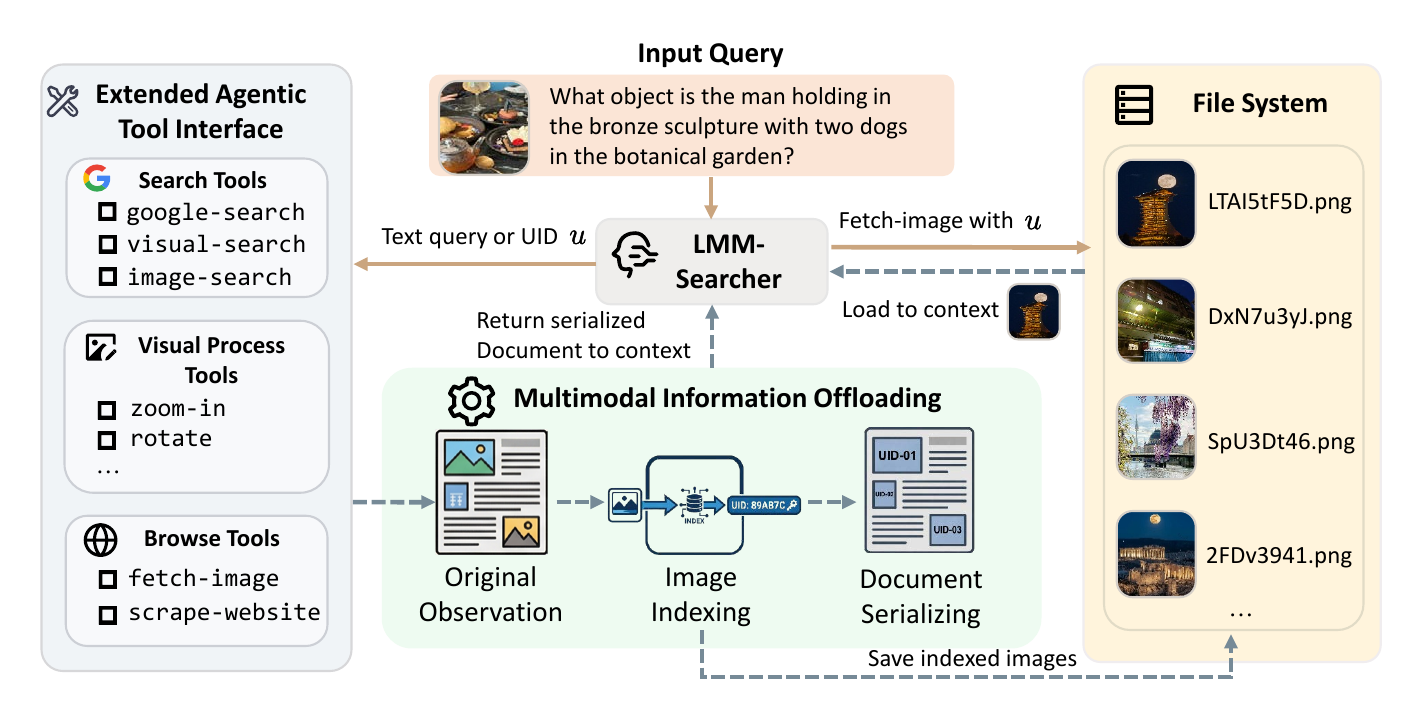}
    \caption{An illustration of LMM-Searcher. For simplicity, we employ simple strings as uids in this figure. In real implementation, we use URLs. This figure serves solely as a functional demonstration and does not represent any actual search results.}
    \label{fig:framework}
\end{figure}


We first propose a context management mechanism that combines file-based multimodal representation with an extended agentic tool interface. The core motivation behind this design is to decouple perception from reasoning. Specifically, while visual perception is inherently ``heavy'', long-horizon planning requires a ``lightweight'' context to prevent token explosion and noise accumulation across multi-turn interactions. Guided by this insight, instead of directly inserting raw multimodal content into the model context, our mechanism stores all visual assets (images in this work) in an external file system and references them through lightweight textual identifiers (UIDs). Based on this design, we equip the agent with specialized tools that actively retrieve and process relevant visual content on demand. This mechanism enables long-horizon interaction for multimodal search agents while preserving fine-grained perceptual capability and avoiding excessive context consumption.

\subsection{File-based Multimodal Data Management}


Throughout the multimodal search process, each image returned by the environment is persistently stored in an external file system. To guarantee that the agent can precisely locate the target image via its UID on demand and subsequently load it into the context, it is essential to establish a strict one-to-one mapping between UIDs and images in the file system.
Formally, let $\mathcal{I}$ denote the high-dimensional visual space and $\mathcal{U}$ denote the space of lightweight textual identifiers~(UID). The file system defines a persistent mapping function $f: \mathcal{I} \rightarrow \mathcal{U}$, such that each retrieved visual asset $i \in \mathcal{I}$ is uniquely associated with a UID $u = f(i)$. Figure~\ref{fig:webpage_case} shows how webpage content is presented to the agent in practice.

\begin{figure}[htbp]
    \centering
    \begin{tcolorbox}[colback=blue!5!white,colframe=blue!55!black,width=0.98\textwidth]
    
    In *Percy Jackson And The Olympians* Season 2 Episode 8 (the finale):
    
    - Percy Jackson is riding in a blue Prius with Sally and his friends.
    
    - The Pegasus Blackjack appears, landing in front of the blue Prius.
    \\
    \\
    \#\#\# Images:
    
    - \textbf{Image URL}: \url{https://images.squarespace-cdn.com/content/v1/5fbc4a62c2150e62cfcb09aa/1769036046809-0WFJWM5XTN0KMM9AZIO1/PercyJacksonSeason2Episode8.jpg}
    
      - \textbf{Caption}: Annabeth, Percy, and Grover back at Camp Half-Blood
    
    - \textbf{Image URL}: \url{https://images.squarespace-cdn.com/content/v1/5fbc4a62c2150e62cfcb09aa/79058544-8368-476a-b2e6-49939bb6737b/Percy-Jackson-Season-2-Episode-8-Thalia%2B1280.jpg}
      
      - \textbf{Caption}: Thalia with her spear
    
    - \textbf{Image URL}: \url{https://images.squarespace-cdn.com/content/v1/5fbc4a62c2150e62cfcb09aa/fd2497a0-85f0-4531-92b3-93210cacf068/Percy-Jackson-Season-2-Episode-8-Poseidon-and-Percy%2B1280.jpg}
      
      - \textbf{Caption}: Poseidon visits Percy in a dream.
    \end{tcolorbox}
    \caption{The webpage returned to the agent. Its content is reorganized into a structured representation, where textual information is summarized into key bullet points and visual elements are converted into image–caption pairs, where the images are replaced with their URLs.}
    \label{fig:webpage_case}
\end{figure}

Through this proxy representation $u$, visual content is converted into a lightweight textual form that can be efficiently maintained in context. When fine-grained visual inspection is required, the agent can actively retrieve the corresponding image through a dedicated tool. To reduce storage overhead, if an image already exists in an external file system (\eg, on the internet) with a valid identifier (\eg, a URL), we directly reuse the existing UID rather than assigning a new one.

\subsection{Extended Agentic Tool Interface}


Previous multimodal search frameworks~\cite{chu2026redsearcher,huang2026vision} integrate various search-related tools. However, these tools often adhere to an ``eager loading'' design paradigm, as they are designed to load images into the model context immediately upon retrieval, leading to a rapid expansion of the context window. To enable long-horizon multimodal search under the proposed file-based representation, we redesign conventional tools to operate over UID-based visual references. Our tool design is grounded in the principle of progressive loading. The designed tool interface includes three categories of tools: \emph{search tools} are responsible for internet searches, \emph{browse tools} handle specific web page content extraction and visual perception, while \emph{visual processing tools} are utilized for editing and finer-grained perception of the extracted images. These three categories of tools collectively form a coarse-to-fine perception funnel. We present a more detailed tool description below:

\paragraph{Search Tools.} 
We integrate open-domain search tools built upon existing search engines (\eg, Serper) as the entry point for cross-modal, multi-hop reasoning. This suite of tools includes \texttt{google\_search}, which accepts textual queries and retrieves relevant documents; \texttt{image\_search}, which takes textual input and returns related images; and \texttt{visual\_search}, which uses an input image to identify visually similar results. These search tools return a set of retrieved items, including textual snippets, image links, thumbnails, and corresponding webpage URLs.

\paragraph{Browse Tools.} We introduce two tools for accessing detailed content from web pages and images. The first, \texttt{scrape\_website}, retrieves and summarizes webpage content. Given a query, a summarization model produces a textual summary while extracting and storing all image URLs from the page. The second, \texttt{fetch\_image}, is designed for active visual perception. Acting as a bridge between the UID space $\mathcal{U}$ and the visual space $\mathcal{I}$, this tool retrieves the image $i$ corresponding to a given UID $u$ from the external file system and provides it to the model for detailed inspection.


\paragraph{Visual Processing Tools.} To support fine-grained visual reasoning, we incorporate an image processing tool (\ie, \texttt{zoom\_in}). Let $g$ denote a visual transformation. Given an input UID $u_{\text{\text{in}}}$, the tool applies the transformation to the underlying image, producing a new visual asset $i_{\text{new}} = g(f^{-1}(u_{\text{in}}))$. The resulting image is then uploaded to the file system and assigned a new identifier $u_{\text{new}} = f(i_{\text{new}})$. Because such operations involve active perception and generate a focused visual result, both $i_{\text{new}}$ and its corresponding UID $u_{\text{new}}$ are inserted into the context simultaneously.

\subsection{Long-horizon Multimodal Search Workflow}

Building upon the file-based data management and the extended tool interface, our agent executes a long-horizon search workflow. This design simulates the human paradigm of information acquisition: we do not maintain high-resolution visual details of every retrieved document in our memory; instead, we remember where the information is and progressively load it when needed. Specifically, during a search task, when the agent invokes \textbf{search tools} or \textbf{browse tools}, the raw output from the environment is an interleaved document $\mathcal{D}$ containing both text and raw images. Crucially, before $\mathcal{D}$ enters the agent's context, our framework acts as an intercepting middleware. It automatically indexes all visual items within $\mathcal{D}$, permanently saves them to the file system, and serializes the document by replacing all raw images with their corresponding UIDs. Consequently, the agent only receives a lightweight representation of the search results. By fully proxying visual content with UIDs at this stage, we effectively solve the context explosion problem. The agent can maintain an extensive search history across dozens of turns without suffering from visual token bloat, successfully decoupling lightweight long-horizon reasoning from heavy visual perception.

When the agent identifies a need for fine-grained perception of a specific image mentioned in the text, it autonomously invokes the \texttt{fetch\_image} tool using the UID. Furthermore, if the visual reasoning requires finer perception and manipulation, the agent triggers the \textbf{visual processing tools}. This dynamic interplay ensures that heavy perception occurs strictly on demand. More importantly, this workflow provides a natural reliability guarantee against information loss. Unlike heuristic methods that aggressively discard images~\cite{hong2025deepeyesv2,chu2026redsearcher,huang2026vision}, our framework ensures that no visual asset is ever irrevocably lost. The UID acts as a persistent, low-cost semantic pointer; as long as the UID is retained in the reasoning chain, the agent is guaranteed to trace back to the exact, uncompressed visual evidence in the external file system whenever needed.
\section{Agentic Training for Multimodal Search}

Based on the above design, we aim to equip the model with the capability to utilize this multimodal context management mechanism, and enhance its long-horizon multimodal search capability through agentic training. Specifically, we propose a comprehensive training pipeline that encompasses query synthesis, trajectory data distillation, and model training. A major bottleneck in prior data synthesis efforts~\cite{chu2026redsearcher} is the scarcity of high-quality queries. While existing datasets typically restrict multimodal inputs to the initial search stage (i.e., explicit image queries), they rarely demand multimodal reasoning in subsequent steps, inherently limiting the task complexity and trajectory quality. To overcome this, we synthesize complex queries that require the model to actively read and comprehend multimodal information across webpages throughout the entire search process. Specifically, our pipeline progresses through the synthesis of Visual Question Answering (VQA) pairs from multimodal webpages (Section \ref{sec:multimodal}) and the extension of reasoning chains (Section \ref{sec:multi-hop}). This is followed by agent trajectory synthesis (Section \ref{sec:trajectory}) to generate the final data utilized for model training (Section \ref{sec:training}).

\subsection{Multimodal Webpage Query Synthesis}
\label{sec:multimodal}
To ensure that the synthesized queries strictly require the model to read multimodal webpage information, we select multimedia websites rich in visual content (\eg, news and movie websites) as a starting point to synthesize single-hop visual questions. This process involves webpage content extraction and question synthesis.

\paragraph{Webpage Content Extraction.}
Given a multimodal webpage $W$, we parse it using Jina\footnote{https://jina.ai} and input it into a MLLM to extract the core entity $E$ in the webpage, along with the image $I_E$ related to $E$. The fundamental principle is that $I_E$ must have direct captions or rich context, and $E$ must be a unique, unambiguous entity. Detailed prompts can be found in the Appendix.

\paragraph{Visual Question Synthesis.}
Based on $W$, $E$, and $I_E$, we prompt a MLLM to synthesize a visual question related to $I_E$, with the constraint that the question cannot be answered solely using the textual information in $W$. Then, we prompt a MLLM to synthesize a clue that mentions both $E$ and $I_E$, based on the relationship between them presented in the webpage $W$. Finally, by combining this clue with the visual question, we obtain a single-hop visual question $q_{0}$. The question $q_{0}$ can only be answered when the search agent successfully navigates to this specific webpage, forcing the agent to continuously compare various acquired multimodal information with the given clues throughout the entire search process.

\begin{figure}
    \centering
    \includegraphics[width=1\linewidth]{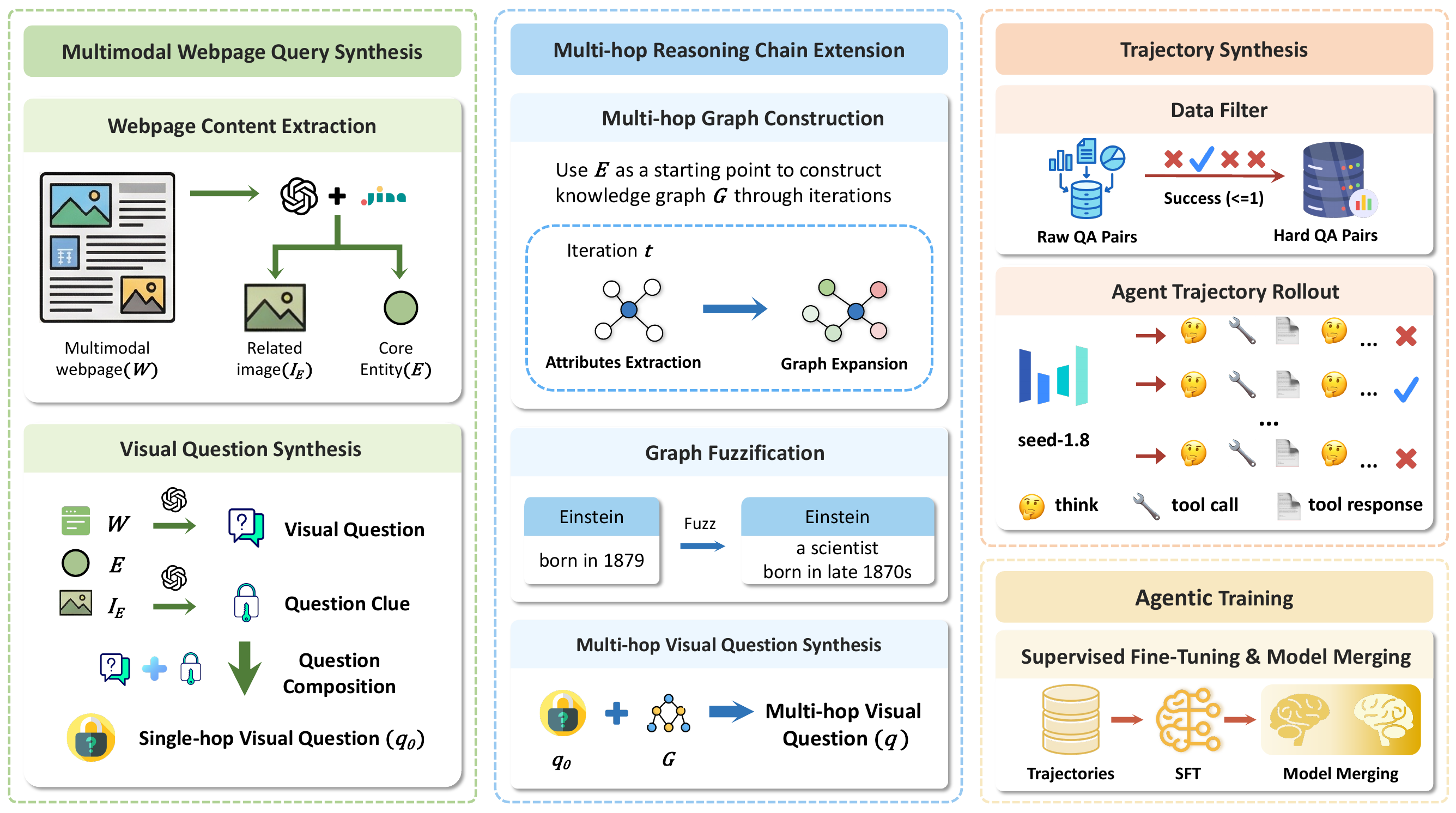}
    \caption{Overview of automated Visual Question-Answer (VQA) data synthesis pipeline. The pipeline constructs multimodal deep search data by synthesizing VQA pairs based on multimodal webpages and subsequently extending the reasoning chain.}
    \label{fig:data_pipeline}
\end{figure}

\subsection{Multi-hop Reasoning Chain Extension}
\label{sec:multi-hop}
To increase the difficulty of the previously obtained queries, we extend the reasoning chain based on $E$. The basic idea is to use $E$ as a starting point to construct a multi-hop knowledge graph denoted as $G$, and then fuzzify the key nodes in $G$ to ultimately generate an extended multi-hop visual question $q$. Specifically, this is divided into the following steps.

\paragraph{Multi-hop Graph Construction.}
We design a workflow to construct a multi-hop knowledge graph iteratively. We define the knowledge graph as a directed graph $G = (\mathcal{V}, \mathcal{E})$, where $\mathcal{V}$ represents the set of nodes and $\mathcal{E} \subseteq \mathcal{V} \times \mathcal{V}$ represents the set of links (edges) between nodes. Each node $v \in \mathcal{V}$ in the graph maintains an expansion state $S(v) \in \{\text{unexpanded}, \text{expanded}\}$. Initially, the graph contains only an isolated root node $v_{\text{root}}=E$, \ie, $G^{(0)} = (\mathcal{V}^{(0)}, \emptyset)$, where $\mathcal{V}^{(0)} = \{v_{\text{root}}\}$ and $S(v_{\text{root}}) = \text{unexpanded}$. In the $t$-th iteration, we select an unexpanded node $v_t \in \mathcal{V}^{(t-1)}$ (\ie, a node with an out-degree of 0) from the current graph $G^{(t-1)}$. Denoting the entity corresponding to this node as $E_t$, we prompt a LLM to extract a set of attributes $\mathcal{R}_t = \{r_t^1, r_t^2, \dots, r_t^m\}$ based on the search results of $E_t$ from knowledge sources (offline databases or internet webpages). Next, based on the density of the graph and the depth of $v_t$, the LLM filters an attribute subset $\hat{\mathcal{R}}_t$ from $\mathcal{R}_t$ ($\hat{\mathcal{R}}_t \subset \mathcal{R}_t$). Each attribute link $r_t^i$ in this subset points to a new, further expandable target entity $u_t^i$. Subsequently, we add these new nodes and directed edges to the graph. The graph expansion process at the $t$-th iteration can be represented as follows:
\begin{equation}
\begin{aligned}
\mathcal{V}^{(t)} &= \mathcal{V}^{(t-1)} \cup \{u_t^i \mid r_t^i \in \hat{\mathcal{R}}_t\} \\
\mathcal{E}^{(t)} &= \mathcal{E}^{(t-1)} \cup \{(v_t, r_t^i, u_t^i) \mid r_t^i \in \hat{\mathcal{R}}_t\}
\end{aligned}
\end{equation}

Upon completion of the update, the state of node $v_t$ transitions to expanded, \ie, $S(v_t) = \text{expanded}$. To increase the complexity of the reasoning chain, we impose strict information obfuscation constraints on the selected attributes: for any single attribute $r_t^i$ in $\hat{\mathcal{R}}_t$, it cannot be used independently to reversely deduce the source entity $v_t$. This constraint of information irreversibility ensures that every newly added edge in the graph plays an indispensable role in the final multi-hop reasoning, thereby preventing the model from taking shortcuts to directly retrieve the answer.

\paragraph{Graph Fuzzification.}
After completing the graph construction, we select entities of leaf nodes and nodes with low in-degrees/out-degrees in the current knowledge graph $G$, and group them into a set of entities to be fuzzified, denoted as $\{F_k\}_{k=1}^n$. These entities lack the necessary edges for reference and thus must be represented in a fuzzified manner. We randomly select an attribute $r_{k}^{j}$ of the target entity $F_k$ that are not used during graph construction and prompt an LLM to fuzzify the node.

\paragraph{Multi-hop Visual Question Synthesis.}
We sample the constructed graph to obtain a subgraph $G'$ containing the core entity $E$. We randomly select a leaf node $E_i$ and replace it with an explicit image containing the corresponding entity. Subsequently, an LLM is prompted to convert the subgraph $G'$ containing the explicit image into a natural language reasoning text that concludes with the core entity $E$. Following this, the LLM inserts this reasoning text into the previously synthesized single-hop visual question $q_{0}$ regarding the core entity $E$, thereby synthesizing a multi-hop visual question $q$.

\subsection{Trajectory Synthesis}
\label{sec:trajectory}
After synthesizing the queries, we further construct agent trajectories for training. To improve query diversity and increase data scale, we additionally incorporate several open-source search-related datasets, including FVQA~\cite{wang2017fvqa}, LiveVQA~\cite{fu2025livevqa}, REDSearcher-Text~\cite{chu2026redsearcher}, and REDSearcher-MM~\cite{chu2026redsearcher}. Following previous studies~\cite{du2025makes,liu2025less}, we implement a preliminary filtering stage for quality control. Specifically, we first use Qwen2.5-VL-7B~\cite{bai2025qwen3} to filter out queries that can be answered correctly without invoking a search engine. The remaining queries are then used for trajectory synthesis. We perform rejection sampling with Seed-1.8~\cite{seed2026seed1}, retaining only trajectories that successfully answer the query within 40 interaction turns under a 64K context length constraint. This process yields 12,736 high-quality training samples in total. The detailed data distribution is summarized in Table~\ref{tab:data_statistic}. Notably, compared with existing datasets, the data synthesized by our pipeline requires substantially longer interaction trajectories, indicating stronger long-horizon search characteristics.

\begin{table}
    \centering
    \begin{tabular}{lccccc}
    \toprule
          Dataset&FVQA&  LiveVQA&  REDSearcher-MM&REDSearcher-Text& Ours \\
    \midrule
          Num. of Samples&2301&  1672&  3366 &3808& 1589 \\
        Avg. Turns& 5.63& 7.16& 13.21 &21.97& 17.26\\
    \bottomrule
    \end{tabular}
    \caption{Dataset statistics, including the number of samples and the average number of tool-use turns per sample.}
    \label{tab:data_statistic}
\end{table}

\subsection{Model Training}
\label{sec:training}
To validate the effectiveness of both our agent framework and the synthesized dataset, we perform multi-turn supervised fine-tuning (SFT) on Qwen3-VL-30B-A3B-Thinking~\cite{bai2025qwen3}, a state-of-the-art open-source multimodal large language model. During training, we mask tool responses when computing the cross-entropy loss, such that the model is optimized only to generate the reasoning process and tool calls. Although we have ensured the high quality of the synthetic data, due to its multimodal nature, multimodal search trajectories often struggle to reach the interaction scale of pure-text search trajectories, which limits the scaling capabilities of the trained model. Inspired by previous studies~\cite{chen2025bring,liu2025vift}, many general scaling capabilities can be transferred between language models and multimodal models through model merging. Consequently, we merge our trained model with MiroThinker-1.7-mini~\cite{team2026mirothinker}, a model that shares the same language model backbone as our target model and has undergone large-scale mid-training and demonstrates strong language-based deep search capabilities. The merging process is specifically applied to the language model parts common to both models. Denoting our trained multimodal model as $\Theta_V$ and MiroThinker-1.7-mini as $\Theta_T$, the final LMM-Searcher-30B model is obtained by parameter interpolation:
\begin{equation}
    \Theta_{\text{final}} = \alpha \cdot \Theta_V +(1-\alpha) \cdot \Theta_T
\end{equation}
We set $\alpha = 0.8$, which preserves most multimodal capabilities while incorporating the strengths of MiroThinker-1.7-mini. A rigorous study of model merging is left for future work.
\section{Experiment}
\begin{table}[t]
    \centering
    \small
    \begin{tabular}{lccccc}
    \toprule
         Model&  MMBC&  MMSearch+ &VisBrowse&  MMSearch&  Avg.\\
    \midrule
 \multicolumn{6}{c}{Direct Answer}\\
    \midrule
         GPT-5& 10.3 &  \textbf{19.1} &26.0&  33.3&  22.2\\
         Seed-1.8& \textbf{13.0}& 8.6 &18.9& 31.0&17.9\\
 Kimi-K2.5& 2.7& 12.6& 18.3& \textbf{47.0}&20.2\\
         Gemini-2.5-Pro& 10.3 &  14.5 &\textbf{27.2}&  39.8&  \textbf{23.0}\\
         Gemini-2.5-Flash& 5.4 &  8.1 &16.0&  30.4&  15.0\\
         Qwen3-VL-30B-A3B-Thinking & 7.1 &  2.7 &  13.0&  17.7  &  10.1\\
    \midrule
         \multicolumn{6}{c}{Agentic Search}\\
    \midrule
         GPT-5&  23.7&  34.8 &35.5&  72.2&  41.6\\
         Seed-1.8& 25.5& \textbf{46.7} &\textbf{58.0}& \textbf{73.2}&\textbf{50.9}\\
         Kimi-K2.5& \textbf{25.9}& 39.2 &50.3& 72.3&46.9\\
         Gemini-2.5-Pro&  12.1&  28.1 &16.0&  66.3&  30.6\\
         Gemini-2.5-Flash&  8.0&  14.0 &14.2&  59.8&  24.0\\
         Qwen3-VL-30B-A3B-Thinking&  9.8&  14.4 &16.0&  62.0&  25.6\\
    \midrule
         \multicolumn{6}{c}{Multimodal Search Agent}\\
    \midrule
         MMSearch-R1-7B&  -&  - &-&  53.8&  -\\
         Webwatcher-7B&  -&  - &-&  49.1&  -\\
         Webwatcher-32B&  -&  - &-&  55.3&  -\\
         DeepEyesV2-7B&  -&   &-&  63.7&  -\\
         Vision-DeepResearch-30B&  -&  28.5 &-&  69.6&  -\\
         REDSearcher-MM-30B&  23.5&  26.6 &-&  \textbf{72.9}&  -\\
         LMM-Searcher-30B&  22.3/\textbf{30.1}$^*$&  32.9/\textbf{34.8}$^*$&42.0/\textbf{48.3}$^*$&  71.0/72.3$^*$&  42.1/\textbf{46.4}*\\
    \bottomrule
    \end{tabular}
    \caption{The performance comparison between our model and baseline methods. The performance marked with $^*$ is evaluated with 100 turns and the context management technique. MMBC shorts for MM-BrowseComp, and MMSearch+ shorts for MMSearch-Plus.}
    \label{tab:main}
\end{table}

\subsection{Experiment Setup}
\paragraph{Evaluation Benchmarks.} We evaluate our model on various challenging visual search benchmarks. The evaluation benchmarks include: MM-BrowseComp~\cite{li2025mm}, VisBrowse-Bench~\cite{zhang2026visbrowse}, MMSearch-Plus~\cite{tao2025mmsearch}, and MMSearch~\cite{jiang2024mmsearch}. Following prior work, we only evaluate on the single-image subset of MMSearch-Plus for fair comparison.

\paragraph{Baselines.}
We consider three categories of baseline methods: direct answer, agent workflow, and multimodal search agents.
\begin{itemize}[leftmargin=1em]
    \item Direct Answer. The model generates responses solely based on its parametric knowledge, without performing any image manipulation or external search.
    \item Agent Workflow. The model is integrated into our agent framework, where it can invoke a suite of tools to assist in answering queries.
    \item Multimodal Search Agents. We compare against existing open-source multimodal agents, including MMSearch-R1~\cite{wu2025mmsearch}, WebWatcher~\cite{geng2025webwatcher}, DeepEyesV2~\cite{hong2025deepeyesv2}, Vision-DeepResearch~\cite{huang2026vision}, and REDSearcher-MM~\cite{chu2026redsearcher}. These methods typically combine perception, reasoning, and external search to address complex multimodal queries.
\end{itemize}

\paragraph{Implementation Details.} We build our framework based on MiroFlow~\cite{miromind2026miroflow}, and utilize it for both trajectory rollout and answer verification. We utilize LLaMA-Factory~\cite{zheng2024llamafactory} as the training framework. We train the model for 3 epochs, with a global batch size of 64 and a learning rate of 1e-5. During evaluation, we set the maximum length as 128K and the maximum number of turns as 30 to make a fair comparison with previous methods. We also report the success rate when extending the turns to 100 and only keep the recent 5 tool call results, which is a context management strategy introduced by DeepSeek-V3.2~\cite{liu2025deepseek}.

\subsection{Main Results}

\subsubsection{Overall Performance}
Table~\ref{tab:main} reports the overall performance comparison across four benchmarks. First, we observe a consistent trend that direct answer methods significantly underperform agent-based approaches, highlighting the necessity of tool use and external search for complex multimodal tasks. Moreover, our LMM-Searcher-30B achieves competitive or superior performance compared with existing multimodal search agents. In particular, LMM-Searcher-30B attains 28.7 on the challenging MMSearch-Plus benchmark, while maintaining strong and comparable performance on MM-BrowseComp and MMSearch. Furthermore, when enabling long-horizon interaction (100 turns) with context management, LMM-Searcher-30B yields consistent improvement across all benchmarks. Notably, it achieves state-of-the-art performance on MM-BrowseComp and MMSearch-Plus, demonstrating its effectiveness in handling extended reasoning and interaction. Overall, these results validate the advantage of our approach in jointly enabling strong search capability, robust multimodal reasoning, and scalable long-horizon interaction.

\subsubsection{Comparison with Other Frameworks}

To validate the generalization capabilities of our context management design, we assess the performance of identical models deployed across both our framework and the previous REDSearcher~\cite{chu2026redsearcher} and Vision-DeepResearch framework~\cite{huang2026vision} on multiple benchmarks. We set the maximum number of interaction turns to 50 to better unleash the framework’s potential and ensure a fair comparison. As shown in \autoref{tab:exp-framework}, our framework consistently improves the average performance across all evaluated models. We discover that models with weaker visual agentic capabilities (\eg, Qwen3-VL-30B-A3B-Thinking) exhibit marginal gains, which suggests that a fixed search-and-look workflow suffices for simpler agents. However, more capable models benefit more significantly under our framework. For example, Seed-1.8 demonstrates an improvement of 13.7\% on MMBC, and 35.7\% on MMSearch-Plus. Furthermore, our approach yields the most substantial improvements on challenging tasks. Specifically, it boosts GPT-5 by 17.6\% on MMSearch-Plus, and Seed-1.8 by 14.3\% on MMBC and 35.7\% on MMSearch-Plus. These results underscore the efficacy of our framework in tackling complex, visual multi-hop problems.


\begin{table}
    \small
    \centering
    \begin{tabular}{llcccc}
    \toprule
    Model & Evaluation Method & MMBC & MMSearch+ & VisBrowse & MMSearch \\
    \midrule
    \multirow{3}{*}{GPT-5} 
    & Direct Answer & 10.3 & 19.1 & 26.0 & 33.3 \\
    & w. Previous Framework & - & 17.2 & - & 63.7 \\
    & w. Our Framework & \textbf{36.5} & \textbf{34.8} & \textbf{35.5} & \textbf{72.2} \\
    \midrule
    \multirow{3}{*}{Seed-1.8} 
    & Direct Answer & 13.0 & 8.6 & 18.9 & 31.0 \\
    & w. Previous Framework & 21.4 & 11.0 & - & 69.7 \\
    & w. Our Framework & \textbf{35.1} & \textbf{46.7} & \textbf{58.0} & \textbf{73.2} \\
    \midrule
    \multirow{3}{*}{\makecell[l]{Qwen3-VL-30B-\\A3B-Thinking}} 
    & Direct Answer & 7.1 & 2.7 & 13.0 & 17.7 \\
    & w. Previous Framework & \textbf{10.7} & 13.6 & - & 53.2 \\
    & w. Our Framework & 9.8 & \textbf{14.4} & \textbf{16.0} & \textbf{62.0} \\
    \bottomrule
    \end{tabular}
    \caption{Performance comparison among different frameworks.}
    \label{tab:exp-framework}
\end{table}

\subsection{Further Analysis}
\begin{figure}
    \centering
    \includegraphics[width=1\linewidth]{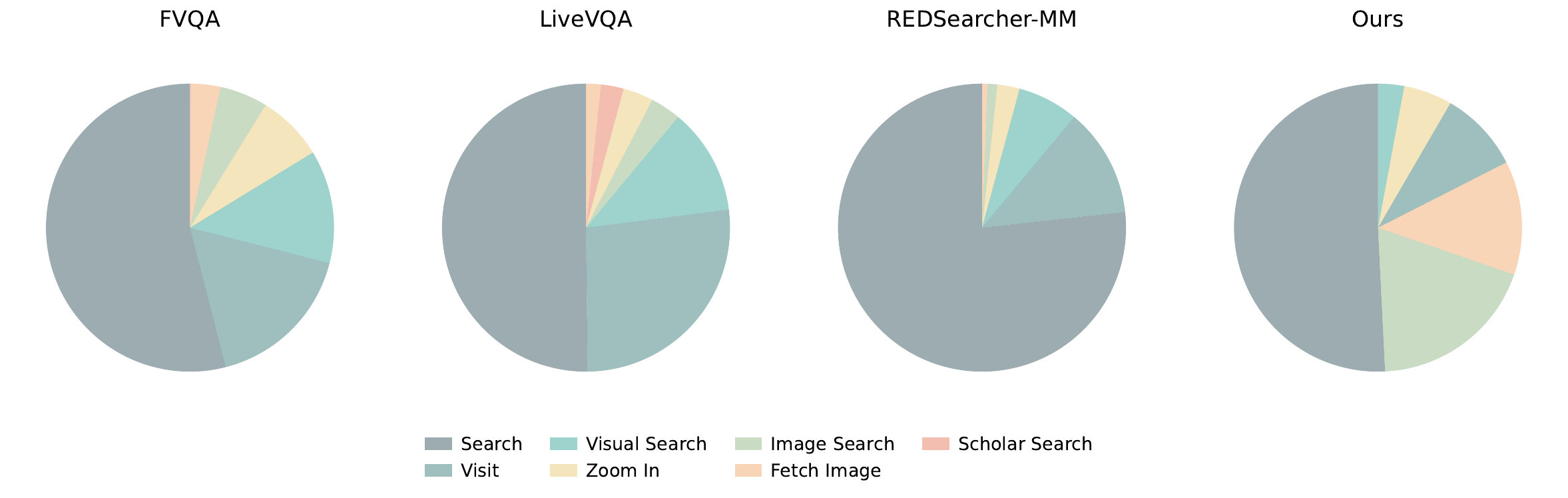}
    \caption{Tool call distribution of the training data. Our synthesized data require diverse types of tool calls and a higher proportion of ``visual search'' and ``fetch image''.}
    \label{fig:tool_dist}
\end{figure}

\paragraph{Tool Distribution.} To demonstrate that our synthesized queries inherently require more intensive multimodal search and browsing, we analyze the distribution of tool calls in agent trajectories and compare it with those induced by existing open-source multimodal search queries. As shown in Figure~\ref{fig:tool_dist}, our queries trigger substantially more ``visual search'' and ``image search''. More importantly, they require significantly more frequent ``fetch image'' steps during problem solving. This indicates that our synthesized queries demand deeper inspection of multimodal content on web pages, rather than relying on superficial retrieval alone.

\paragraph{Interaction Scaling.}


\begin{figure}[htbp]
    \centering
    \begin{subfigure}{0.45\textwidth}
        \centering
        \includegraphics[width=\textwidth]{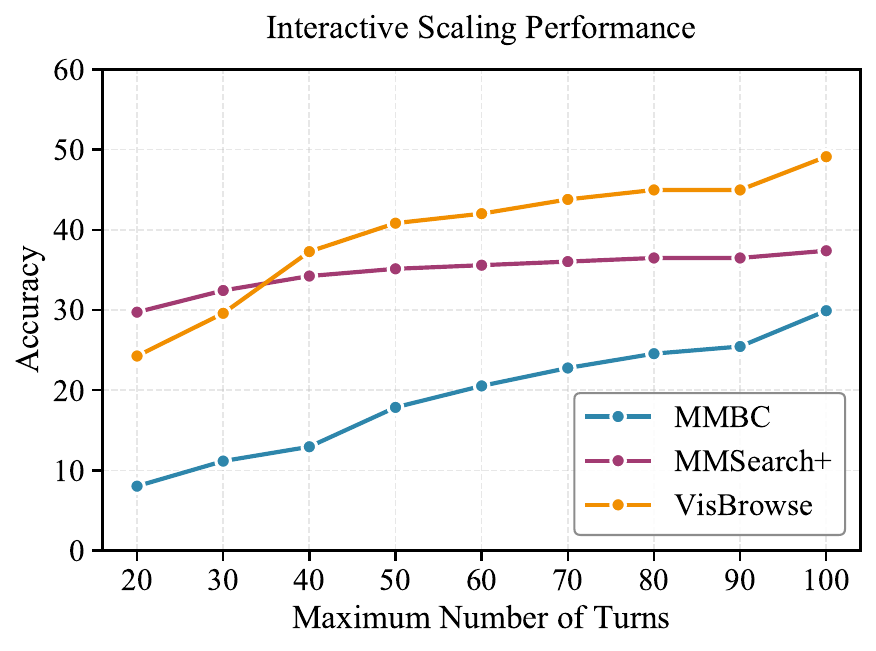}
        \caption{Scaling results on different benchmarks.}
        \label{fig:scaling-benchmark}
    \end{subfigure}
    \hfill
    \begin{subfigure}{0.45\textwidth}
        \centering
        \includegraphics[width=\textwidth]{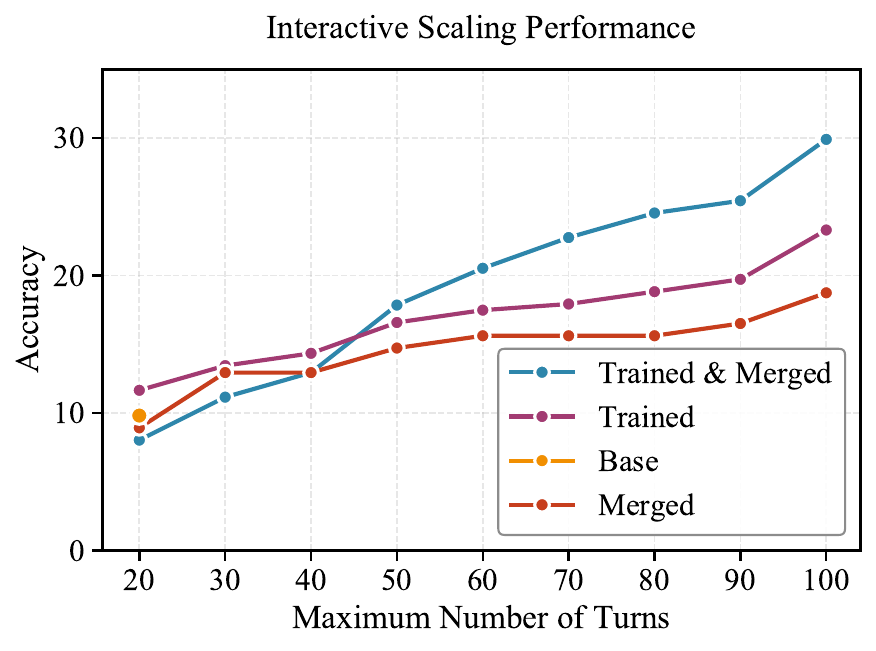}
        \caption{Scaling results on model training and merging.}
        \label{fig:scaling-merge}
    \end{subfigure}
    
    \caption{Interactive scaling results. We evaluate different models with our context management strategy and force the agent to stop at 100 turns. For each interaction-turn threshold on the x-axis, we count a sample as successful if the agent has autonomously terminated and produced the correct answer within that number of turns, and compute the corresponding accuracy accordingly.} 
    \label{fig:scaling}
\end{figure}

To evaluate the interactive scaling behavior of LMM-Searcher, we impose varying turn limits, truncating interactions that exceed the threshold and marking them as incomplete. Our evaluation focuses on two aspects: (1) the scaling performance of LMM-Searcher-30B across diverse benchmarks, and (2) the comparison of different model variants (\ie, base, trained, merged, and final) on MMBC. The results in Figure~\ref{fig:scaling} show that our model consistently benefits from increased interaction turns across all benchmarks, although the magnitude of improvement varies, suggesting different tasks require different reasoning depths. Notably, MMBC and VisBrowse continue to improve even at 100 turns, indicating strong scaling potential for multimodal search. Furthermore, Figure~\ref{fig:scaling-merge} shows that while the base model saturates around 20 turns, both synthetic data training and language model merging substantially enhance its scaling behavior, with their combination yielding further gains. These results demonstrate that our approach effectively transfers long-horizon capabilities from language-based search to multimodal settings.

\paragraph{Data Ablation.} To validate the effectiveness of our data synthesis pipeline, we conduct an ablation study over training data from different sources and modalities. Specifically, we consider open-source multimodal search queries (including FVQA, LiveVQA, and REDSearcher-MM), open-source textual queries (REDSearcher-Text), and our synthesized multimodal search queries. Each dataset is incrementally added on top of the previous training set. After training, we merge the resulting checkpoint with MiroThinker-1.7-mini via model merging. As shown in Table~\ref{tab:data_ablation}, training with only open-source multimodal search queries already leads to substantial improvements in agentic search performance. Incorporating additional textual queries brings gains primarily on the long-horizon benchmark MMBC, but leads to performance degradation on average. This may be because the gains brought by additional textual queries are already largely covered by the language-based search capability inherited through model merging. Further introducing our synthesized queries yields additional improvements on MMBC and VisBrowse, and ultimately achieves the best average performance across all benchmarks.

\begin{table}
    \centering
    \begin{tabular}{lccccc}
    \toprule
         &  MMBC&  MMSearch+&  VisBrowse& MMSearch &Avg.\\
    \midrule
         Qwen3-VL-30B-A3B-Thinking &  9.8&  14.4&  16.0&  62.0&25.6\\
         + Open-source Visual Query&  20.7&  \textbf{33.8}&  39.5&  70.4&41.1\\
         + Open-source Textual Query&  21.6&  32.4&  39.1&  70.3&40.9\\
         + Our Synthesized Query& \textbf{22.3}& 32.9& \textbf{42.0}& \textbf{71.0}&\textbf{42.1}\\
    \bottomrule
    \end{tabular}
    \caption{Data ablation results. Each dataset is incrementally added on top of the previous training set.}
    \label{tab:data_ablation}
\end{table}

\paragraph{Tool Ablation.} A core difference between our approach and previous multimodal deep search frameworks is that we save all multimodal information encountered during the search process as files, allowing the agent to load them flexibly. The primary tool facilitating this operation is ``fetch-image''. To directly validate the effectiveness of this tool, we used Seed-1.8 as the base model and compared the performance with and without it within our current framework. As shown in Table \ref{tab:tool_ablation}, removing the fetch-image tool degrades the performance of Seed-1.8 across all benchmarks. The most significant decline occurs on the VisBrowse benchmark, dropping from 58.0 to 48.5, which indicates that this benchmark heavily relies on acquiring image information from webpages. Conversely, the performance on MMSearch only decreases from 73.2 to 71.0, suggesting that this benchmark primarily relies on search engine results and does not require the agent to visit the webpages.
\begin{table}
    \centering
    \begin{tabular}{lccccc}
    \toprule
         &  MMBC&  MMSearch+&  VisBrowse& MMSearch &Avg.\\
    \midrule
         Seed-1.8&  9.8&  14.4&  16.0&  62.0&25.6\\
         w/ fetch-image tool&  \textbf{35.1}&  \textbf{46.7}&  \textbf{58.0}&  \textbf{73.2}&\textbf{53.3}\\
         wo/ fetch-image tool& 29.5& 43.7& 48.5& 71.0&48.2\\
         $\Delta$ & -5.6& -3.0& -9.5& -2.2&-5.1\\
    \bottomrule
    \end{tabular}
    \caption{Tool ablation results. ``w/ fetch-image'' represents equipping the agent with a full tool interface, while ``wo/ fetch-image tool'' represents removing the fetch-image tool.}
    \label{tab:tool_ablation}
\end{table}
\section{Conclusion}
In this work, we present LMM-Searcher, an open-source multimodal deep search agent capable of
resolving complex multimodal queries. We build a long-horizon multimodal context management mechanism with file-based visual representation and carefully designed agentic visual tools, enabling efficient handling of multimodal content and long-horizon interactions.
Furthermore, we develop a dedicated data synthesis pipeline—including multimodal query synthesis and agent trajectory rollout—to construct a high-quality dataset that substantially improves agent performance through SFT. The
evaluations on four multimodal deep search benchmarks show that LMM-Searcher-30B achieves advanced performance among
open-source search agents. These results demonstrate the effectiveness of our end-to-end approach,
scalable framework design, and data synthesis technique in
advancing multimodal deep search agents.

\clearpage

\bibliographystyle{unsrt}
\bibliography{main}

\clearpage

\appendix
\section{Case Study}
\begin{figure}[hbp]
    \centering
    \includegraphics[width=0.95\textwidth]{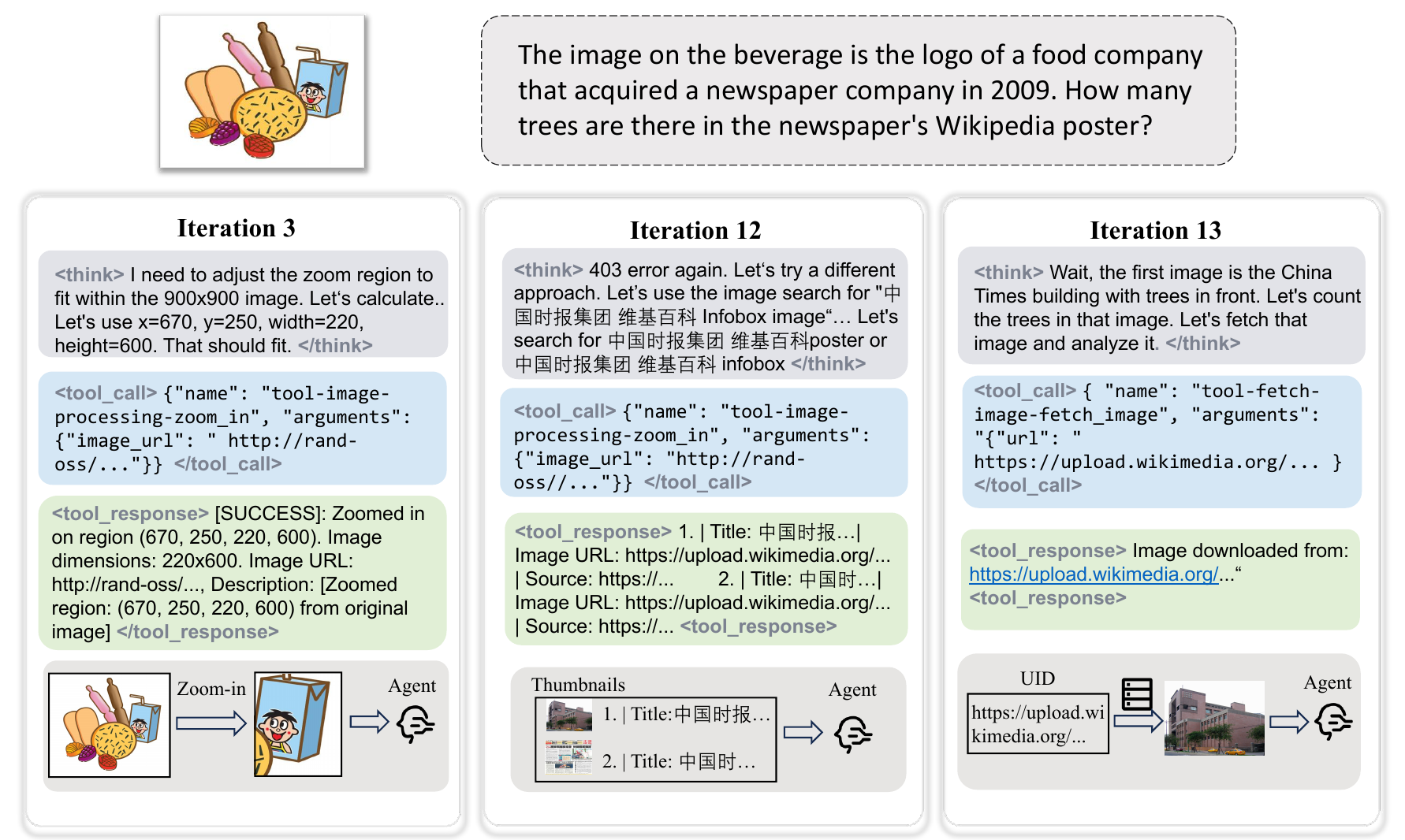}
    \caption{Detailed illustration of key steps within the model's search trajectory.}
    \label{fig:case-study-1}
\end{figure}

To concretely demonstrate the workflow of our framework, we select a case from VisBrowse-Bench and present some detailed step-level illustrations in \autoref{fig:case-study-1}, and its complete reasoning trajectory in \autoref{fig:case-study-2}. As we can observe, our model exhibits the following capabilities: (1) Visual agentic tool use: The model can autonomously select tools to facilitate enhanced perception. In Iteration 3, it successfully delineates the precise zoom-in region, aiding in the identification of the correct brand. (2) Alternating reasoning and perception: From Iteration 5 to Iteration 13, through continuous reasoning, the model alternately employs text search and visual search tools to locate relevant images. In Iteration 14, the model proactively loads the intermediate search images into the context for perception, successfully deriving the correct answer. (3) Reflection: In Iterations 3 and 12, drawing upon previous failed tool calls, the model reflects to formulate new reasoning paths, thereby advancing the search process. These capabilities effectively enhance the model's cross-modal reasoning performance, enabling it to scale search turns within a limited context.

\begin{table*}[htbp]
    \centering
    \renewcommand{\arraystretch}{1.2} 
    \setlength{\tabcolsep}{8pt} 
    \begin{tabular}{p{14cm}}
        \centering

        \begin{tcolorbox}[breakable, colback=gray!5, colframe=gray!40, sharp corners, width=\linewidth, boxrule=0.5mm, title={Examples of our LMM-Searcher on VisBrowse-Bench (Part 1/4)}]
        \small
        \textbf{Data Source:} VisBrowse-Bench 
        
        \textbf{Problem: }
        The image on the beverage is the logo of a food company that acquired a newspaper company in 2009. How many trees are there in the newspaper's Wikipedia poster?
        Image URL: [http://rand-oss//2F\%2F9bXW.jpg]  Description: [Original input image]

        \textbf{Ground Truth:} 5

        \vspace{1.5em}
        \noindent\colorbox{gray!30}{
        \makebox[\dimexpr\linewidth-2\fboxsep\relax][c]{\textbf{Iteration 1}}
        }

        \texttt{<\textbf{think}>} First step is to identify the food company logo in the beverage image. Let's use visual search on the beverage (the blue carton with the cartoon character). \texttt{<\textbf{/think}>} 

        \texttt{<\textbf{tool\_call}>} \{"name": "tool-google-search-visual\_search", "arguments": \{"image\_url": "http://rand-oss//2FP2F9bXW"\}\} \texttt{<\textbf{/tool\_call}>} 
        
        \texttt{<\textbf{tool\_response}>} 1. | Title: Giay tu cong bo san pham FROZEN COLA | Image URL: https://wantwant.vn/vnt..| Source: https://wantwant.vn/cn/gi.. 2. | Title: Facts For Kids- France | Image URL: https://www.oocities.org/... | Source: https://www.oocities.org/... \texttt{<\textbf{/tool\_response}>} 

        \vspace{1.5em}
        \noindent\colorbox{gray!30}{
        \makebox[\dimexpr\linewidth-2\fboxsep\relax][c]{\textbf{Iteration 2}}
        }

        \texttt{<\textbf{think}>} The visual search didn't directly identify the logo. Let's try zooming in on the blue beverage carton part of the image, since that's the logo area. Let's find the coordinates of the blue carton. Looking at the image, the blue carton is on the right side. Let's define the region: x around 670, y around 250, width around 300, height around 400. Let's zoom in on that area. \texttt{<\textbf{/think}>} 

        \texttt{<\textbf{tool\_call}>} \{ "name": "tool-image-processing-zoom\_in", "arguments": \{"image\_url": "http://rand-oss/..."\} \} \texttt{<\textbf{/tool\_call}>} 
        
        \texttt{<\textbf{tool\_response}>} [\{"error": "Your cropped region extends beyond image bounds (image size: 900x900, your cropped region: (670+300)x(250+400))"\}] \texttt{<\textbf{/tool\_response}>} 

        \vspace{1.5em}
        \noindent\colorbox{gray!30}{
        \makebox[\dimexpr\linewidth-2\fboxsep\relax][c]{\textbf{Iteration 3}}
        }

        \texttt{<\textbf{think}>} I need to adjust the zoom region to fit within the 900x900 image. Let's calculate: max x + width <=900, max y + height <=900. So if x=670, width can be up to 230 (900-670). If y=250, height can be up to 650 (900-250). Let's use x=670, y=250, width=220, height=600. That should fit. \texttt{<\textbf{/think}>} 

        \texttt{<\textbf{tool\_call}>} \{"name": "tool-image-processing-zoom\_in", "arguments": \{"image\_url": "http://rand-oss/..."\}\} \texttt{<\textbf{/tool\_call}>} 

        \texttt{<\textbf{tool\_response}>} [SUCCESS]: Zoomed in on region (670, 250, 220, 600). Image dimensions: 220x600. Image URL: http://rand-oss/..., Description: [Zoomed region: (670, 250, 220, 600) from original image] \texttt{<\textbf{/tool\_response}>} 

        \end{tcolorbox}
    \end{tabular}
\caption{A case study illustrating the complete search trajectory of the model.}
\label{fig:case-study-2}
\end{table*}

\begin{table*}[htbp]
    \centering
    \renewcommand{\arraystretch}{1.2} 
    \setlength{\tabcolsep}{8pt} 
    \begin{tabular}{p{14cm}}
        \centering
        \begin{tcolorbox}[breakable, colback=gray!5, colframe=gray!40, sharp corners, width=\linewidth, boxrule=0.5mm, title={Examples of our LMM-Searcher on VisBrowse-Bench (Part 2/4)}]
        \small
        
        \vspace{0.5em}
        \noindent\colorbox{gray!30}{
        \makebox[\dimexpr\linewidth-2\fboxsep\relax][c]{\textbf{Iteration 4}}
        }

        \texttt{<\textbf{think}>} Now I can see the logo clearly: it's Want Want's mascot (the boy with the red cheeks). Want Want is a food company. The question says this company acquired a newspaper company in 2009. Let me confirm: Want Want acquired a newspaper? Let me search for "Want Want acquired newspaper 2009". \texttt{<\textbf{/think}>} 

        \texttt{<\textbf{tool\_call}>} \{"name": "tool-google-search-google\_search", "arguments": "\{"q": "Want Want acquired newspaper 2009"\}"\} \texttt{<\textbf{/tool\_call}>} 
        
        \texttt{<\textbf{tool\_response}>} 1. "title": "China Times Group is sold to Want Want", "link": "https://www.taipeitimes.com/News/biz/...", "snippet": "It owns the Chinese-language newspapers China Times..." 2. "title": "China Times Joins a Snack-food Empire|Industry|2009-03-05", "link": "https://english.cw.com.tw/article/...", "snippet": "A month after Want Want Holdings chairman Tsai Eng-meng ..." ...... \texttt{<\textbf{/tool\_response}>} 

        \vspace{1.5em}
        \noindent\colorbox{gray!30}{
        \makebox[\dimexpr\linewidth-2\fboxsep\relax][c]{\textbf{Iteration 5}}
        }

        \texttt{<\textbf{think}>} 
        So the newspaper company is China Times Group (\begin{CJK*}{UTF8}{gbsn}中时集团\end{CJK*}), acquired by Want Want in 2008 (the deal closed in 2009? The news says completed in March 2009). Now, the question asks: "How many trees are there in the newspaper's Wikipedia poster?" ... Let me search for the Wikipedia page of the newspaper acquired, then check the image (poster) for tree count.
        First, let's get the Wikipedia page for China Times Group. Let me search for "China Times Group Wikipedia" or "\begin{CJK*}{UTF8}{gbsn}旺旺中时媒体集团\end{CJK*} Wikipedia" (since it's in Chinese).
        \texttt{<\textbf{/think}>} 

        \texttt{<\textbf{tool\_call}>} \{ "name": "tool-google-search-google\_search", "arguments": "\{"q": "Want Want acquired newspaper 2009"\}" \} \texttt{<\textbf{/tool\_call}>} 

        \texttt{<\textbf{tool\_response}>} 
        1. "title": "\begin{CJK*}{UTF8}{gbsn}旺旺中时媒体集团 - 维基百科\end{CJK*}", "link": "https://zh.wikipedia.org/zh-cn/..."
        2. "title": "\begin{CJK*}{UTF8}{gbsn}可靠来源/常见/旺旺中时媒体集团 - 维基百科\end{CJK*}", "link": "https://zh.wikipedia.org/zh-cn/..."
        ......
        \texttt{<\textbf{/tool\_response}>} 

        \vspace{1.5em}
        \noindent\colorbox{gray!30}{
        \makebox[\dimexpr\linewidth-2\fboxsep\relax][c]{\textbf{Iteration 6}}
        }

        \texttt{<\textbf{think}>} 
        Now, let's open the Wikipedia page for \begin{CJK*}{UTF8}{gbsn}旺旺中时媒体集团\end{CJK*} (https://zh.wikipedia.org/zh-cn/\begin{CJK*}{UTF8}{gbsn}旺旺中时媒体集团\end{CJK*}). We need to find the "poster" (infobox image) and count the number of trees in it. Let's use the scrape\_website tool to get the content of this page, ... Let's call the scrape\_and\_extract\_info tool with the URL and ask to count the trees in the Wikipedia poster (the infobox image).
        \texttt{<\textbf{/think}>} 

        \texttt{<\textbf{tool\_call}>} \{ "name": "jina\_scrape\_llm\_summary-scrape\_and\_extract\_info", 
        "arguments": "\{"url": "https://zh.wikipedia.org/zh-cn/...", "info\_to\_extract": "Count the number of trees in the Wikipedia poster of \begin{CJK*}{UTF8}{gbsn}旺旺中时媒体集团\end{CJK*}"\}"
        \} \texttt{<\textbf{/tool\_call}>} 

        \texttt{<\textbf{tool\_response}>} 
        \{"success": false, "url": "https://zh.wikipedia.org/zh-cn/\begin{CJK*}{UTF8}{gbsn}旺旺中时媒体集团\end{CJK*}", "extracted\_info": "", "error": "Scraping failed (both Jina and Python): Python Scrape: Unexpected error occurred: Client error '403 Forbidden' for url 'https://zh.wikipedia.org/zh-cn/..."\}
        \texttt{<\textbf{/tool\_response}>} 

        \vspace{1.5em}
        \noindent\colorbox{gray!30}{
        \makebox[\dimexpr\linewidth-2\fboxsep\relax][c]{\textbf{Iteration 7}}
        }

        \texttt{<\textbf{think}>} 
        403 Forbidden error. Let's try a different approach. Let's first find the infobox image URL for this Wikipedia page. ... Let's search for "\begin{CJK*}{UTF8}{gbsn}旺旺中时媒体集团 维基百科 图片\end{CJK*}".
        \texttt{<\textbf{/think}>} 

        \texttt{<\textbf{tool\_call}>} 
        \{ "name": "tool-google-search-image\_search", "arguments": "\{"q": "\begin{CJK*}{UTF8}{gbsn}旺旺中时媒体集团 维基百科 图片\end{CJK*}"\}" \}
        \texttt{<\textbf{/tool\_call}>} 

        \texttt{<\textbf{tool\_response}>} 
        1. | Title: \begin{CJK*}{UTF8}{gbsn}旺旺中时媒体集团- 维基百科，自由的百科全书\end{CJK*} | Image URL: https://upload.wikimedia.org/wikipedia/zh/thumb/0/05/... | Source: https://zh.wikipedia.org/zh-tw/...
        2. | Title: \begin{CJK*}{UTF8}{gbsn}旺旺集团- 维基百科，自由的百科全书\end{CJK*} | Image URL: https://upload.wikimedia.org/wikipedia/commons/7/74/... | Source: https://zh.wikipedia.org/zh-tw/...
        \texttt{<\textbf{/tool\_response}>} 

        \end{tcolorbox}
    \end{tabular}
\end{table*}

\begin{table*}[htbp]
    \centering
    \renewcommand{\arraystretch}{1.2} 
    \setlength{\tabcolsep}{8pt} 
    \begin{tabular}{p{14cm}}
        \centering
        \begin{tcolorbox}[breakable, colback=gray!5, colframe=gray!40, sharp corners, width=\linewidth, boxrule=0.5mm, title={Examples of our LMM-Searcher on VisBrowse-Bench (Part 3/4)}]
        \small

        \vspace{0.5em}
        \noindent\colorbox{gray!30}{
        \makebox[\dimexpr\linewidth-2\fboxsep\relax][c]{\textbf{Iteration 8}}
        }

        \texttt{<\textbf{think}>} 
        Wait, the first image is the logo of Want Want China Times Media Group. Let's check that image. The logo has the Want Want kid, "\begin{CJK*}{UTF8}{gbsn}中时集团\end{CJK*}" in red squares, and "CHINA TIMES GROUP". Is there a tree in this logo? Let's fetch this image and analyze it.
        \texttt{<\textbf{/think}>} 

        \texttt{<\textbf{tool\_call}>} \{ "name": "tool-fetch-image-fetch\_image", "arguments": "\{"url": "https://upload.wikimedia.org/wikipedia/zh/thumb/0/05/..."\}" \} \texttt{<\textbf{/tool\_call}>} 

        \texttt{<\textbf{tool\_response}>} Image downloaded from: https://upload.wikimedia.org/wikipedia/.. \texttt{<\textbf{/tool\_response}>} 

        \vspace{1.5em}
        \noindent\colorbox{gray!30}{
        \makebox[\dimexpr\linewidth-2\fboxsep\relax][c]{\textbf{Iteration 9}}
        }

        \texttt{<\textbf{think}>} 
        Looking at this logo, I don't see any trees. Wait, maybe the "newspaper's Wikipedia poster" refers to the China Times (China Times Group) poster, not the media group's poster. Let's check the Wikipedia page for China Times Group (\begin{CJK*}{UTF8}{gbsn}中国时报集团\end{CJK*}). Let's search for "\begin{CJK*}{UTF8}{gbsn}中国时报集团 维基百科\end{CJK*}" and get its poster. Alternatively, maybe the "poster" is the Wikipedia infobox image for China Times Group. Let's search for the China Times Group Wikipedia page. Let me call the google search for "\begin{CJK*}{UTF8}{gbsn}中国时报集团 维基百科\end{CJK*}".
        \texttt{<\textbf{/think}>} 

        \texttt{<\textbf{tool\_call}>} 
        \{ "name": "tool-google-search-google\_search", "arguments": "\{"q": "\begin{CJK*}{UTF8}{gbsn}中国时报集团 维基百科\end{CJK*}"\}" \}
        \texttt{<\textbf{/tool\_call}>} 

        \texttt{<\textbf{tool\_response}>} 
        1. "title": "\begin{CJK*}{UTF8}{gbsn}讨论:中国时报集团- 维基百科，自由的百科全书\end{CJK*}", "link": "https://zh.wikipedia.org/zh-my/..."
        2. "title": "\begin{CJK*}{UTF8}{gbsn}中时新闻网\end{CJK*}", "link": "https://www.wikiwand.com/zh-hant/..."
        \texttt{<\textbf{/tool\_response}>} 

        \vspace{1.5em}
        \noindent\colorbox{gray!30}{
        \makebox[\dimexpr\linewidth-2\fboxsep\relax][c]{\textbf{Iteration 10}}
        }

        \texttt{<\textbf{think}>} 
        Wait, maybe the Wikipedia page for China Times Group is under a different name? Let me try "\begin{CJK*}{UTF8}{gbsn}中时集团 维基百科\end{CJK*}".
        \texttt{<\textbf{/think}>} 

        \texttt{<\textbf{tool\_call}>} 
        \{ "name": "tool-google-search-google\_search", "arguments": "\{"q": "\begin{CJK*}{UTF8}{gbsn}中时集团 维基百科\end{CJK*}"\}" \}
        \texttt{<\textbf{/tool\_call}>} 

        \texttt{<\textbf{tool\_response}>} 
        1. "title": "\begin{CJK*}{UTF8}{gbsn}中国时报集团- 维基百科，自由的百科全书\end{CJK*}", "link": "https://zh.wikipedia.org/zh-cn/..."
        2. "title": "\begin{CJK*}{UTF8}{gbsn}旺旺中时媒体集团 - 维基百科\end{CJK*}", "link": "https://zh.wikipedia.org/zh-cn/..."
        \texttt{<\textbf{/tool\_response}>} 

        \vspace{1.5em}
        \noindent\colorbox{gray!30}{
        \makebox[\dimexpr\linewidth-2\fboxsep\relax][c]{\textbf{Iteration 11}}
        }

        \texttt{<\textbf{think}>} 
        Okay, so the newspaper company is China Times Group (\begin{CJK*}{UTF8}{gbsn}中国时报集团\end{CJK*}), which was acquired by Want Want in 2008-2009. Now, the question is asking for the number of trees in the newspaper's Wikipedia poster. Let's find the Wikipedia poster (infobox image) for China Times Group. Let's scrape the Wikipedia page for China Times Group and look for the infobox image. Let's use the jina\_scrape tool on that page, asking to find the infobox image and count the trees in it.
        \texttt{<\textbf{/think}>} 

        \texttt{<\textbf{tool\_call}>} 
        \{ "name": "jina\_scrape\_llm\_summary-scrape\_and\_extract\_info", "arguments": "\{"url": "https://zh.wikipedia.org/zh-cn/...", "info\_to\_extract": "Find the infobox image (poster) on the Wikipedia page for \begin{CJK*}{UTF8}{gbsn}中国时报集团\end{CJK*} and count the number of trees in it."\}" \}
        \texttt{<\textbf{/tool\_call}>} 

        \texttt{<\textbf{tool\_response}>} 
        \{"success": false, "url": "https://zh.wikipedia.org/zh-cn/...", "extracted\_info": "", "error": "Scraping failed (both Jina and Python): Python Scrape: Unexpected error occurred: Client error '403 Forbidden' for url 'https://zh.wikipedia.org/zh-cn/..."\}
        \texttt{<\textbf{/tool\_response}>} 

        \end{tcolorbox}
    \end{tabular}
\label{tab:case5_part3}
\end{table*}

\begin{table*}[htbp]
    \centering
    \renewcommand{\arraystretch}{1.2} 
    \setlength{\tabcolsep}{8pt} 
    \begin{tabular}{p{14cm}}
        \centering
        \begin{tcolorbox}[breakable, colback=gray!5, colframe=gray!40, sharp corners, width=\linewidth, boxrule=0.5mm, title={Examples of our LMM-Searcher on VisBrowse-Bench (Part 4/4)}]
        \small

        \vspace{0.5em}
        \noindent\colorbox{gray!30}{
        \makebox[\dimexpr\linewidth-2\fboxsep\relax][c]{\textbf{Iteration 12}}
        }

        \texttt{<\textbf{think}>} 
        403 error again. Let's try a different approach. Let's use the image search for "\begin{CJK*}{UTF8}{gbsn}中国时报集团 维基百科\end{CJK*} Infobox image". Or maybe directly find the infobox image URL via Google search image. Let's search for "\begin{CJK*}{UTF8}{gbsn}中国时报集团 维基百科\end{CJK*} poster" or "\begin{CJK*}{UTF8}{gbsn}中国时报集团 维基百科\end{CJK*} infobox".
        \texttt{<\textbf{/think}>} 

        \texttt{<\textbf{tool\_call}>} 
        \{ "name": "tool-google-search-image\_search", "arguments": "\{"q": "\begin{CJK*}{UTF8}{gbsn}中国时报集团 维基百科\end{CJK*} infobox"\}" \}
        \texttt{<\textbf{/tool\_call}>} 

        \texttt{<\textbf{tool\_response}>} 
        1. | Title: \begin{CJK*}{UTF8}{gbsn}中国时报集团- 维基百科，自由的百科全书\end{CJK*} | Image URL: https://upload.wikimedia.org/wikipedia/commons/e/ee/... | Source: https://zh.wikipedia.org/zh-tw/...
        2. | Title: \begin{CJK*}{UTF8}{gbsn}中国时报- 维基百科，自由的百科全书\end{CJK*} | Image URL: https://upload.wikimedia.org/wikipedia/zh/9/9b/... | Source: https://zh.wikipedia.org/zh-cn/...
        \texttt{<\textbf{/tool\_response}>} 

        \vspace{1.5em}
        \noindent\colorbox{gray!30}{
        \makebox[\dimexpr\linewidth-2\fboxsep\relax][c]{\textbf{Iteration 13}}
        }

        \texttt{<\textbf{think}>} Wait, the first image is the China Times building with trees in front. Let's count the trees in that image. Let's fetch that image and analyze it. \texttt{<\textbf{/think}>} 

        \texttt{<\textbf{tool\_call}>} \{ "name": "tool-fetch-image-fetch\_image", "arguments": "\{"url": "https://upload.wikimedia.org/wikipedia/commons/e/ee/..."\}" \} \texttt{<\textbf{/tool\_call}>} 

        \texttt{<\textbf{tool\_response}>} Image downloaded from: https://upload.wikimedia.org/..." \texttt{<\textbf{/tool\_response}>} 

        \vspace{1.5em}
        \noindent\colorbox{gray!30}{
        \makebox[\dimexpr\linewidth-2\fboxsep\relax][c]{\textbf{Iteration 14}}
        }

        \texttt{<\textbf{think}>} 
        Now let's count the trees in this image. Let's see: in front of the building, there are several trees. Let's list them: left side: one tree, then two more, .... Wait, but let me confirm once more. Let me list each tree with coordinates: 1. (x: 20, y: 800) - small tree  2. (x: 200, y: 850) - tree near pole  3. (x: 450, y: 880) - tree near first taxi  4. (x: 550, y: 890) - tree near second taxi  5. (x: 900, y: 850) - large tree on right Yes, that's 5 trees. So the answer is 5.
        \texttt{<\textbf{/think}>} 
        
        The food company is Want Want, which acquired the **China Times Group** (\begin{CJK*}{UTF8}{gbsn}中时媒体集团\end{CJK*}) in 2009. The Wikipedia infobox poster for this newspaper shows **5 distinct trees** in front of its headquarters building."

        \vspace{1.5em}
        \noindent\colorbox{gray!30}{
        \makebox[\dimexpr\linewidth-2\fboxsep\relax][c]{\textbf{Iteration 15}}
        }

        \texttt{<\textbf{User}>} Summarize the above conversation, and output the FINAL ANSWER to the original question. If a clear answer has already been provided earlier in the conversation, do not rethink or recalculate it — simply extract that answer and reformat it to match the required format below.  If a definitive answer could not be determined, make a well-informed educated guess based on the conversation...

        \texttt{<\textbf{think}>} Okay, let's summarize the conversation and extract the final answer. First, we identified the logo on the beverage as Want Want's mascot. Then we found that Want Want acquired China Times Group in 2009. We looked at the Wikipedia poster (infobox image) for China Times Group, which shows the headquarters building with trees. Counting the distinct trees, we found 5.The original question asks for the number of trees in the newspaper's Wikipedia poster. So the answer is boxed\{5\}. \texttt{<\textbf{/think}>} 

        \end{tcolorbox}
    \end{tabular}
\end{table*}

\end{document}